\pdfoutput=1
\documentclass[11pt]{article}

\usepackage{ACL2023}

\usepackage{times}
\usepackage{latexsym}
\usepackage[T1]{fontenc} % For proper rendering and hyphenation
\usepackage[utf8]{inputenc} % This assumes your files are encoded as UTF8
\usepackage{microtype} % Improves typography and layout
\usepackage{inconsolata} % Improves aesthetics of typewriter font
\usepackage{tabularx} % 导言区
\usepackage{amsmath}
\usepackage{amssymb}
\usepackage{amsfonts} % For blackboard math symbols
\usepackage{nicefrac} % For compact symbols like 1/2

\usepackage{graphicx}
\usepackage{booktabs} % For professional-quality tables
\usepackage{multirow}
\usepackage{subfigure}
\usepackage{threeparttable}
\usepackage{wrapfig}
\usepackage{booktabs}
\usepackage{multirow}
\usepackage{adjustbox}
\usepackage{stfloats}

\usepackage[table]{xcolor} % 灰色分块
\usepackage{siunitx}       % 数字对齐
\usepackage{algorithm}
\usepackage{algpseudocode} % Modern and flexible package for pseudocode

\usepackage{url}
\usepackage{xcolor}
\usepackage{natbib}
\usepackage{comment}
\usepackage{lipsum} % For placeholder text with wrapfig, can likely be removed
\usepackage{mdframed}
\definecolor{theoremcolor}{rgb}{0.97, 0.97, 0.97}
\definecolor{examplecolor}{rgb}{1, 1, 1.0}
\mdfsetup{
    innertopmargin=8pt,
    innerbottommargin=8pt,
    leftmargin=4pt,
    rightmargin=4pt,
    backgroundcolor=theoremcolor,
    linewidth=0pt,
}
\newmdtheoremenv{theorem}{Theorem}
\newmdtheoremenv{definition}{Definition}
\newmdtheoremenv{proposition}{Proposition}
\newmdtheoremenv{corollary}{Corollary}
\newmdtheoremenv{lemma}{Lemma}
\newmdtheoremenv{remark}{Remark}
\newmdtheoremenv[linewidth=0pt,innerleftmargin=0pt,innerrightmargin=0pt,backgroundcolor=examplecolor]{example}{Example}

\title{Martingale Foresight Sampling: A Principled Approach to Inference-Time LLM Decoding}

\author{
 Huayu Li\textsuperscript{1} \quad
 \And
 ZhengXiao He\textsuperscript{1} \quad
 \AND
 Siyuan Tian\textsuperscript{2} \quad
 \And
 Jinghao Wen\textsuperscript{3} \quad
 \And
 Ao Li\textsuperscript{1}\thanks{\, denotes corresponding author.} \\
 \textsuperscript{1} University of Arizona \\
 \textsuperscript{2} Microsoft Research Asia
 \textsuperscript{3} Villanova University
 }

\author{
Huayu Li\textsuperscript{1}\thanks{\, equal contribution, names are ordered alphabetically by first name.} \quad
ZhengXiao He\textsuperscript{1*} \quad
Siyuan Tian\textsuperscript{2} \quad
Jinghao Wen\textsuperscript{3} \quad \\
\bf{
Ao Li\textsuperscript{1}\thanks{\, denotes corresponding author.} \quad
}\\
\textsuperscript{1} University of Arizona \quad
\textsuperscript{2} Microsoft Research \quad
\textsuperscript{3} Villanova University \quad \\
\texttt{\{hl459,zhengxiaohe,aoli1\}@arizona.edu} \:
\\
}

\begin{document}
\maketitle
\begin{abstract}
Standard autoregressive decoding in large language models (LLMs) is inherently short-sighted, often failing to find globally optimal reasoning paths due to its token-by-token generation process. While inference-time strategies like foresight sampling attempt to mitigate this by simulating future steps, they typically rely on ad-hoc heuristics for valuing paths and pruning the search space. This paper introduces Martingale Foresight Sampling (MFS), a principled framework that reformulates LLM decoding as a problem of identifying an optimal stochastic process. By modeling the quality of a reasoning path as a stochastic process, we leverage Martingale theory to design a theoretically-grounded algorithm. Our approach replaces heuristic mechanisms with principles from probability theory: step valuation is derived from the Doob Decomposition Theorem to measure a path's predictable advantage, path selection uses Optional Stopping Theory for principled pruning of suboptimal candidates, and an adaptive stopping rule based on the Martingale Convergence Theorem terminates exploration once a path's quality has provably converged. Experiments on six reasoning benchmarks demonstrate that MFS surpasses state-of-the-art methods in accuracy while significantly improving computational efficiency. Code will be released at \url{https://github.com/miraclehetech/EACL2026-Martingale-Foresight-Sampling}.
\end{abstract}

\section{Introduction}
Large Language Models (LLMs) have demonstrated a remarkable capacity for complex reasoning, largely through their ability to generate step-by-step chains of thought \citep{wei2022chain}. This technique mimics human deliberation, breaking down daunting problems into a sequence of manageable steps. By generating text one token at a time, the model is fundamentally myopic. It makes a series of locally optimal bets, choosing the most probable next word without any true awareness of the long-term consequences. This process often leads it down paths that, while promising at first, ultimately culminate in logical fallacies or factual dead ends, forcing a stark choice: settle for fast but fragile reasoning, or invest heavily in more computationally intensive methods to guide the model toward a globally coherent solution.

The first wave of solutions approached this challenge as a classical search problem. Pioneering frameworks such as Tree-of-Thought (ToT) \citep{yao2023tree} explicitly cast the reasoning process as navigation through a vast decision tree. By exploring multiple branches simultaneously, evaluating intermediate thoughts and backtracking from unpromising paths, these methods can uncover robust solutions that elude simple greedy decoding. However, this exhaustive exploration comes at a steep and often prohibitive cost. The exponential branching factor of language turns the reasoning space into an immense labyrinth, and traversing it demands computational resources that scale poorly, rendering these methods impractical for many applications.

To strike a more sustainable balance between performance and efficiency, a more recent and promising paradigm has emerged: \emph{Foresight Sampling} \citep{manon}. Instead of blindly exploring the entire search space, this strategy intelligently peers into the future, using short, simulated rollouts to estimate the value of a potential step. The state-of-the-art implementation of this idea, \textit{$\phi$-Decoding} \citep{xu2025phi}, has shown considerable success. It operationalizes foresight by combining two intuitive but heuristic scores: an Advantage score to reward immediate progress and a clustering-based Alignment score to measure consensus among different possible futures. This approach represents a significant leap forward, offering a more targeted and efficient exploration of the solution space.

However, beneath this empirical success lies a fragile foundation. The architecture of these advanced decoders is built upon a collection of ad-hoc design choices and clever rules of thumb. Valuation functions, dynamic pruning criteria, and alignment metrics are the products of sophisticated craft rather than a rigorous scientific principle. This reliance on heuristics creates a ceiling on progress. Such mechanisms can be brittle and may not generalize across different tasks or models and lack the theoretical guaranties required for truly reliable systems. They leave a crucial gap in our understanding of why they work and more importantly, how they might be systematically improved. This state of affairs raises a critical question for the field: \emph{Can we move beyond heuristic-based design and formulate a principled, theoretically-grounded framework for foresight-based decoding?}

In this paper, we respond definitively with an affirmative answer by proposing a fundamental paradigm shift. We reframe LLM decoding entirely: It is not a search problem to be navigated with heuristics but rather the challenge of identifying an optimal stochastic process. For this, we turn to the elegant and powerful mathematics of \emph{Martingale theory}, the study of fair games, and stochastic processes. This new perspective allows us to model the evolving quality of a reasoning path over time. Within this framework, an optimal path emerges as one that behaves like a \emph{submartingale}, a process whose value is expected on average to increase at every step. Our goal is thus transformed into finding the reasoning path that represents the most favorable game.

This theoretical lens empowers us to systematically derive principled mechanisms that directly replace the ad-hoc components of prior work. Our main contributions are: \textbf{(1) A New Theoretical Framework}, we reframe LLM decoding from a heuristic search into the problem of identifying an optimal stochastic process. The goal is to find reasoning paths that behave like submartingales, ensuring their quality is expected to increase at each step; \textbf{(2) Principled Step Valuation,}  we derive a single, principled score for each reasoning step from the Doob Decomposition Theorem. This predictable advantage metric replaces the heuristic Advantage + Alignment combination used in prior work; \textbf{(3) Optimal Path Pruning,} we leverage Optional Stopping Theory to create a principled method for pruning suboptimal paths during a beam search. This ensures that computational resources are focused on the most promising candidates; \textbf{(4) Adaptive Stopping Rule,} guided by the Martingale Convergence Theorem, we design a dynamic rule that stops the expensive deliberation process as soon as a path's quality has provably converged, preventing wasted computation; \textbf{(5) The MFS Algorithm,} we present the Martingale Foresight Sampling (MFS) algorithm, a concrete and efficient implementation of our theoretical framework that puts these principled concepts into practice.

By replacing fragile rules with mathematical principles, MFS represents a significant step toward building more robust, efficient, and understandable reasoning algorithms for LLMs. Our work not only introduces a high-performing new algorithm, but also provides a foundational framework for future research on principled decoding.

\begin{figure*}[t] % t=top, h=here, b=bottom, p=page of floats
    \centering
    \includegraphics[width=.75\linewidth]{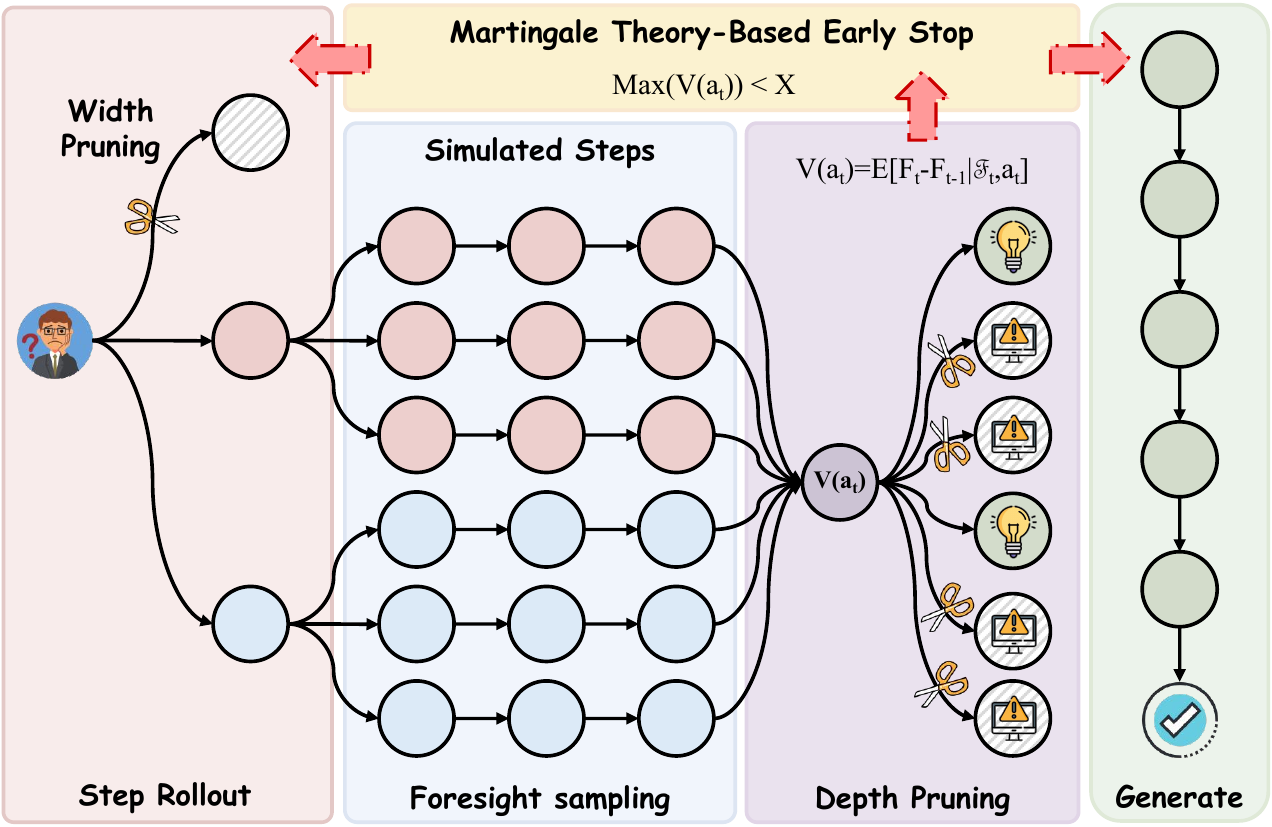}
    \caption{The overall framework of our principled MFS algorithm. We visualized the foresight decoding within one step. For clear visualization, we set the beam size as 3 and rollout number as 3.}
    \label{fig:main-framework}
\end{figure*}

\section{Preliminaries}

\subsection{Foresight Sampling for LLM Decoding}

In standard auto-regressive language generation, the selection of the current token, $a_t$, is conditioned solely on the input query, $x$, and the sequence of previously generated tokens, $a_{<t}$. This can be expressed as sampling from the conditional probability distribution learned by the LLM, $p_{\theta}$:
\begin{equation}
    \hat{a}_{t}\sim p_{\theta}(a_{t}|x,a_{<t})
\end{equation}

This process is inherently myopic, as the choice of $a_t$ is made without any awareness of its downstream consequences on the full reasoning path. Foresight sampling aims to mitigate this limitation by incorporating an estimate of future outcomes into the decoding process. Generation is conditioned not only on the past $a_{<t}$, but also on simulated future token sequences $a_{>t}$.

This concept is formalized by introducing a step value function, $R(x, a_{\le t}, a_{>t})$, which scores a candidate token based on the quality of the future it is expected to produce. The decoding objective then becomes sampling from a modified distribution that incorporates this value:
\begin{equation}
    \hat{a}_{t}\sim p_{\theta}(a_{t}|x,a_{<t})\exp[R(x,a_{\le t},a_{>t})/\tau]
\end{equation}
Here, $\tau$ is a temperature hyperparameter that controls the sharpness of the distribution. The critical challenge lies in defining the function $R$. A key quantity for this is the \textbf{foresight probability}, which estimates the quality of a path at step $t$ by calculating the model's confidence in its simulated future:
\begin{equation}
    F_{t}=p_{\theta}(a_{>t}|x,a_{t},a_{<t})
\end{equation}
Current state-of-the-art methods like \textit{$\phi$-Decoding} construct the reward function $R$ using heuristics based on this foresight probability, such as combining the change in $F_t$ with scores from clustering different future paths. Our work replaces these heuristics with the principled framework of Martingale theory.

\subsection{Martingale Theory}
We now introduce the core concepts from Martingale theory that form the mathematical foundation of our approach.

\begin{definition}[Filtration and Adapted Process]
A \textbf{filtration} is a sequence of increasing $\sigma$-algebras $(\mathcal{F}_n)_{n \ge 0}$ where $\mathcal{F}_n \subseteq \mathcal{F}_{n+1} \subseteq \mathcal{F}$ for all $n$. A stochastic process $(X_n)_{n \ge 0}$ is said to be \textbf{adapted} to the filtration if for every $n$, the random variable $X_n$ is $\mathcal{F}_n$-measurable.
\end{definition}
A filtration $(\mathcal{F}_n)$ represents the accumulation of information over time. At each step $n$, $\mathcal{F}_n$ contains all the information known up to that point. A process being adapted simply means that its value at time $n$, $X_n$, can be determined using only the information available at that time.

\begin{definition}[Martingale]
An adapted process $(M_n)_{n \ge 0}$ is a \textbf{martingale} with respect to a filtration $(\mathcal{F}_n)_{n \ge 0}$ if it has finite mean ($\mathbb{E}|M_n| < \infty$) and satisfies the condition:
\begin{equation}
    \mathbb{E}[M_{n+1} | \mathcal{F}_n] = M_n \quad \text{a.s.}, \quad \forall n \geq 0
\end{equation}

If the equality is replaced by `$\geq$` it is a \textbf{submartingale} ($\mathbb{E}[M_{n+1} | \mathcal{F}_n] \ge M_n$), and if by `$\leq$` it is a \textbf{supermartingale} ($\mathbb{E}[M_{n+1} | \mathcal{F}_n] \le M_n$).
\end{definition}
A martingale models a fair game, where the expected value of your holdings at the next step, given all past outcomes, is exactly your current holdings. A submartingale models a favorable game, where you expect your holdings to increase (or stay the same), while a supermartingale models an unfavorable one.

\begin{theorem}[Doob Decomposition]
Any adapted $L^1$ process $(X_n)_{n \ge 0}$ can be uniquely decomposed into a martingale $(M_n)_{n \ge 0}$ and a predictable process $(A_n)_{n \ge 0}$ such that $X_n = X_0 + M_n + A_n$. The increment of the predictable process is given by:

\begin{equation}
    A_k - A_{k-1} = \mathbb{E}[X_k - X_{k-1} | \mathcal{F}_{k-1}]
\end{equation}

\end{theorem}
This powerful theorem allows us to dissect any stochastic process into two parts: its underlying, knowable trend (the \textbf{predictable process} $A_n$) and its purely random, unpredictable fluctuations (the \textbf{martingale} $M_n$). A process is a submartingale if and only if its predictable part, or drift is non-decreasing. This decomposition is central to our method for valuing reasoning steps.

\begin{theorem}[Doob's Forward Convergence]
A supermartingale that is bounded in $L^1$ converges almost surely to a finite limit.
\end{theorem}
This theorem implies that a process modeling a favorable game (our submartingale quality process) that cannot grow infinitely and is bounded from above must eventually settle down. If the process stops showing a clear upward trend and begins to behave like a martingale, it has likely converged to its limit. Continuing the process beyond this point is inefficient, as no significant improvement is expected. This provides a principled basis for an early-stopping mechanism.

\begin{definition}[Stopping Time]
A random variable $T$ is a \textbf{stopping time} if the event $\{T \le n\}$ is $\mathcal{F}_n$-measurable for all $n$.
\end{definition}
A stopping time is a rule for deciding when to stop a process, where the decision at any time $n$ must be based solely on the information available up to that point. You cannot peek into the future to make your decision. This concept is fundamental to making principled decisions about pruning search paths during decoding. The properties of martingales at stopping times are described by the \textbf{Optional Stopping Theorem}, which justifies that a path that has fallen significantly behind is unlikely to recover, making its pruning a sound decision.

\section{Problem Formulation: Decoding as a Stochastic Process}

We formally model the LLM decoding process within the Martingale framework, transforming the task of generating a sequence of tokens into the problem of identifying an optimal stochastic process. This allows us to leverage the powerful analytical tools of probability theory to guide the decoding procedure. The core components of our model are defined as follows.
\begin{itemize}
    \item \textbf{Filtration ($\mathcal{F}_t$)}: The information revealed up to step $t$ is the sequence of generated tokens $a_{<t}$. This generates the filtration $\mathcal{F}_t = \sigma(a_0, a_1, \dots, a_{t-1})$.\\
    \textbf{Interpretation}: The filtration $\mathcal{F}_t$ is the mathematical formalization of the history available to the language model at each step of the auto-regressive process. Any decision made at step $t$, including the choice of the next token a $a_t$, must be a function of the information contained within $\mathcal{F}_t$.
    \item \textbf{Quality Process ($F_t$)}: For a given reasoning path, we define its quality at step $t$ using the foresight probability $F_t$. The sequence of these quality estimates, $\{F_t\}_{t \ge 0}$, forms a stochastic process adapted to the filtration $\mathcal{F}_t$.\\
    \textbf{Interpretation}: By treating the sequence of foresight probabilities as a stochastic process, we can analyze its trajectory over time. This process, $F_t$, quantifies the model's confidence in the future success of its current reasoning path. Our goal is to steer this process in a favorable direction.
\end{itemize}
With this formulation, the objective of inference-time optimization can be reframed from a heuristic search into a clear, mathematical goal: \emph{To find and extend a reasoning path such that its corresponding quality process $F_t$ is a submartingale with the steepest possible ascent.}

This objective implies that at each step t, we must select the next token $a_t$ that ensures the quality process adheres to the submartingale property:
\begin{equation}
    \mathbb{E}[F_{t+1}|\mathcal{F}_{t}]\geq F_{t}
\end{equation}
This condition states that the expected quality at the next step, given the information we will have then, should be greater than or equal to our current quality. We are explicitly searching for favorable game paths, the reasoning paths that are expected to improve.

Furthermore, we aim for the steepest possible ascent. This transforms the selection of the next token into a well-defined optimization problem. Given a set of candidate tokens for the current step, we seek the token $a^*_t$ that maximizes the expected increase in quality:
\begin{equation}
    a_t^{*} = \underset{a_t}{\arg\max}\;
          \mathbb{E}\!\left[\,F_{t+1} - F_t \;\middle|\; \mathcal{F}_{t}\right]
\end{equation}
This expected increase is precisely the predictable advantage or the drift of the quality process, which we will later show can be estimated using the Doob Decomposition. By maximizing this quantity, we greedily select the path that is most like a submartingale at every step. This principled objective replaces the heuristic valuation functions of prior work and provides a solid theoretical foundation for the decoding mechanisms introduced in the following sections.

\section{Martingale Foresight Sampling}
Having formulated LLM decoding as the search for an optimal stochastic process, we now construct the MFS algorithm. Any practical decoder must answer three fundamental questions: (1) How should we value each potential step? (2) How do we efficiently manage multiple candidate paths in a search? (3) When should the resource-intensive deliberation process end? Our framework allows us to answer each of these questions not with heuristics but with principled mechanisms derived directly from Martingale theory. The overall framework is illustrated in Fig.~\ref{fig:main-framework}. Once the Martingale-theory-based early stopping criterion is activated, only the remaining candidate paths are retained. These surviving paths are executed until completion and their predictions are aggregated via majority voting to produce the final answer. This design ensures that no additional output tokens are generated while also mitigating cases where the model fails to elicit an explicit answer in the output.

\subsection{Principled Step Value via Doob Decomposition}
The first and most critical question for any guided decoder is how to score a potential next token. Our problem formulation seeks to find the path of reasoning whose quality process, $F_t$, behaves as a submartingale with the steepest possible ascent. This requires a value function that accurately reflects this objective. 

The Doob Decomposition Theorem, introduced in our preliminaries, provides a perfect theoretical tool for this task. The theorem states that any adapted process can be uniquely separated into its predictable trend (or drift) and its unpredictable fluctuations. A process is a submartingale if and only if its predictable trend is nondecreasing. To find the submartingale with the steepest ascent, we must therefore choose the step that maximizes this predictable, positive trend.

This insight directly yields our step value function. The value of a candidate step $a_t$ is its predictable advantage, defined as the expected increase of the predictable component of the quality process:
\begin{equation}
\label{eq:predictable_advantage}
V(a_t) = \mathbb{E}[F_t - F_{t-1} | \mathcal{F}_{t}, a_t]
\end{equation}
Although this theoretical quantity is an expectation and cannot be computed exactly, it can be efficiently estimated via Monte Carlo simulation using foresight rollouts. For each candidate token $a_t$, we generate $N$ simulated future paths (rollouts) and average their resulting qualities to approximate the expectation.

This single, theoretically-derived value, $V(a_t)$, replaces the ad-hoc combination of Advantage and Alignment scores used in prior work like $\phi$-Decoding. It provides a unified and principled metric that is directly tied to our objective of finding a path with a positive trend.

\subsubsection{Empirical Validation of Step Values}
We further validate the step value by analyzing token-level log-probability
statistics on correct vs.\ incorrect reasoning paths (LLaMA-3.1-8B-Instruct, ReClor). The
results reveal a clear contrast: correct paths exhibit substantially higher
predictable advantage ($-0.458$ vs.\ $-1.027$) and markedly lower variance
($0.249$ vs.\ $0.994$). These trends confirm that successful reasoning behaves as
a stable, positive-drift process, whereas incorrect reasoning is volatile. This
empirically supports our submartingale formulation and justifies pruning
high-variance trajectories in MFS.

\subsection{Optimal Path Selection via Optional Stopping Theory}
To effectively explore the vast reasoning space, multiple candidate paths must be maintained in parallel, a technique commonly known as a beam search. This introduces the second key question: How do we decide when to prune a suboptimal path to focus computational resources on more promising candidates?

We frame this decision as a stopping-time problem, as defined in our preliminaries. For each competing path $i$ in our beam, we are deciding the optimal time to terminate further investment in its exploration. To formalize this, we define the deficit process
\begin{equation}
D^i_t = F^{\text{best}}_t - F^i_t,
\end{equation}
which measures how far the path $i$ lags behind the current best-scoring path at step $t$. If the best path follows a strong submartingale, while path $i$ does not, the deficit $D^i_t$ itself behaves as a positive drift submartingale, implying that the gap is likely to widen as reasoning progresses.

This motivates a principled pruning rule based on an adaptive stopping time. Specifically, we define
\begin{equation}
T_{\text{prune}}^i = \inf { t \ge 1 \mid D_t^i \ge c_{\text{prune}}(t) },
\end{equation}
where the pruning threshold $c_{\text{prune}}(t)$ is recomputed at every step $t$ according to the empirical score distribution across the beam. Concretely, we use
\begin{equation}
c_{\text{prune}}(t) = \mu_F(t) + \lambda_1\times\sigma_F(t),
\end{equation}
where $\mu_F(t)$ and $\sigma_F(t)$ denote the mean and standard deviation of candidate scores at step $t$, and $\lambda_1$ is a hyperparameter controlling the sensitivity of pruning. This dynamic formulation ensures that a path is pruned only if its deficit is statistically significant relative to the contemporaneous distribution of scores.

The soundness of this mechanism is supported by Doob’s Submartingale Inequality. Once a deficit $D^i_t$ surpasses $c_{\text{prune}}(t)$, the probability that path $i$ will ever catch up to the best path is provably bounded and decays rapidly as the gap widens. Thus, pruning is not heuristic but a theoretically principled application of optional stopping theory, providing a robust, in-search alternative to ad-hoc pre-filtering of candidates.

\begin{figure*}[t]
    \centering
    \includegraphics[width=0.9\linewidth]{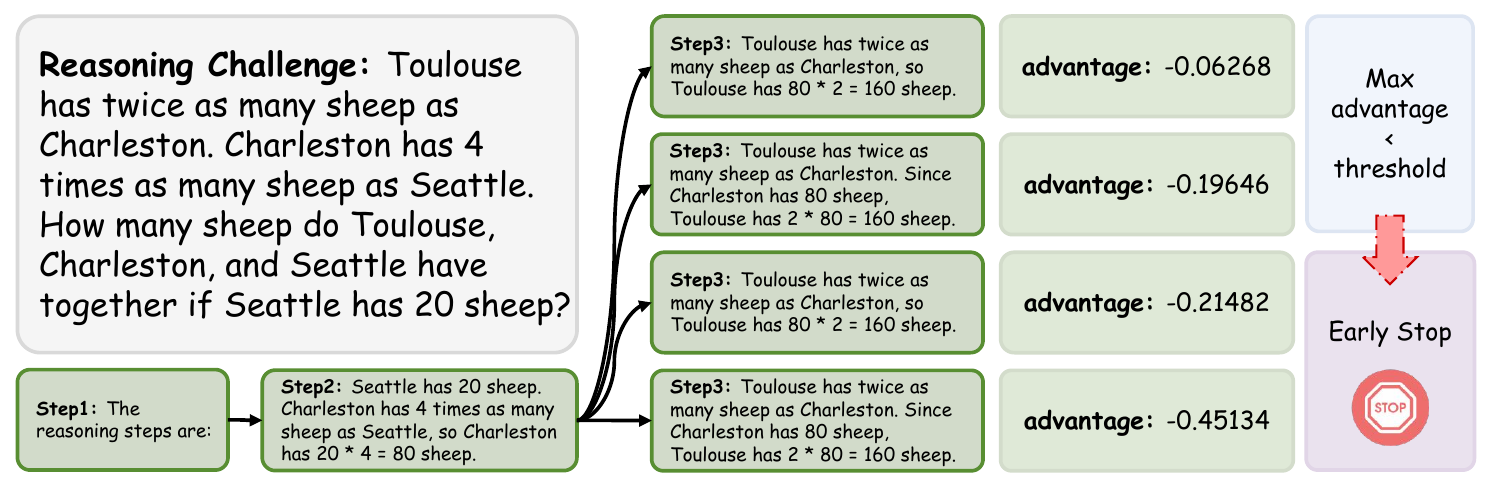}
    \caption{
    Illustration of martingale-based early stopping on a GSM8K example. 
    The deficit process triggers a stopping time once the candidate path’s gap exceeds the adaptive threshold, 
    pruning unpromising trajectories and reducing unnecessary token generation. 
    This principled criterion ensures efficient beam management while preserving solution quality. 
    }
    \label{fig:Concrete Example}
\end{figure*}

\subsection{Adaptive Pruning via Martingale Convergence}
Finally, a crucial aspect of efficiency is knowing when to stop the expensive, step-by-step foresight process altogether and revert to standard auto-regressive generation to complete the answer. Continuing sampling for too long when the process is already stable is wasteful, while stopping too early harms performance. This raises our third question: what is the optimal moment to cease deliberation? 

The Martingale Convergence Theorem provides a clear answer. It tells us that a bounded submartingale (our favorable game) must eventually converge to a finite limit. The practical implication is that if our quality process $F_t$ ceases to show a clear upward trend and begins to behave like a martingale (a fair game), it has likely plateaued. Continuing the foresight process offers no further expected improvement and would waste computational resources.

This theoretical insight translates directly into an adaptive, in-depth pruning rule. Deliberation should cease when the predictable advantage of even the best possible next step falls below a small threshold, $\epsilon_{\text{stop}}\geq 0$,
\begin{equation}
\label{eq:stopping_condition}
\max_{a_t} V(a_t) \le \epsilon_{\text{stop}}
\end{equation}
This condition formally indicates the departure from a strict submartingale behavior, thereby yielding a principled and adaptive stopping criterion. Unlike heuristic cut-offs such as fixed search depths or clustering-based rules, our criterion dynamically adjusts to problem complexity. To further illustrate the early-stopping mechanism, we present Fig.~\ref{fig:Concrete Example}. As shown, once the model has confidently derived the intermediate quantities for Toulouse, Charleston, and Seattle, the final step of summation becomes trivial and deterministically resolved. At this point, additional sampling is redundant; the framework halts exploration, substantially reducing unnecessary token consumption while preserving predictive accuracy.

\section{Experiments}
\begin{table*}[htbp]
\centering
\renewcommand{\arraystretch}{0.9}
\setlength{\tabcolsep}{3pt}
\footnotesize
\begin{tabularx}{\textwidth}{>{\raggedright\arraybackslash}p{3cm}|*{6}{>{\centering\arraybackslash}X}|>{\centering\arraybackslash}p{1.8cm}|>{\centering\arraybackslash}p{2.2cm}}
\toprule
\textbf{Models} & \textbf{GSM8K} & \textbf{Math-500} & \textbf{GPQA} &
\textbf{ReClor} & \textbf{LogiQA} & \textbf{ARC-c} & \textbf{Avg.} & \textbf{FLOPS} \\
\midrule
\rowcolor{gray!12}
\multicolumn{9}{c}{\textbf{LLaMA3.1-8B-Instruct}}\\
Auto-Regressive (CoT)  & 70.28 & 31.00 & 26.56 & 49.40 & 33.33 & 58.91 & 44.91 & $1.34 \times 10^{16}$ \\
Tree-of-Thoughts       & 75.74 & 31.60 & 31.25 & 59.00 & 45.93 & 80.72 & 54.04 & $7.03 \times 10^{17}$ \\
MCTS                   & 80.44 & 34.40 & 24.11 & 61.40 & 42.70 & 79.95 & 53.83 & $1.79 \times 10^{18}$ \\
Guided Decoding        & 75.51 & 31.20 & 30.58 & 60.20 & 43.47 & 81.74 & 53.78 & $6.54 \times 10^{17}$ \\
Predictive Decoding    & 81.43 & 34.00 & 31.03 & 64.00 & 46.70 & 84.56 & 56.95 & $6.89 \times 10^{17}$ \\
\addlinespace[2pt]
$\phi$-Decoding        & 86.58 & 38.20 & 34.60 & 64.00 & 48.39 & 85.41 & 59.53 & $6.43 \times 10^{17}$ \\
Ours                   & \textbf{87.64} & \textbf{38.20} & \textbf{34.90} & \textbf{65.20} & \textbf{48.61} & \textbf{86.73} & \textbf{60.21} & $\mathbf{4.38 \times 10^{17}}$ \\
% token usage: 9738262 & 11761109 & 12782496 & 5697127 & 6159596 & 8611737 & -- & -- \\
\midrule
\rowcolor{gray!12}
\multicolumn{9}{c}{\textbf{Mistral-v0.3-7B-Instruct}}\\
Auto-Regressive (CoT)  & 49.05 & 12.20 & 23.88 & 52.20 & 37.02 & 69.54 & 40.65 & $0.81 \times 10^{16}$ \\
Tree-of-Thoughts       & 53.90 & 10.80 & 26.34 & 55.60 & 41.63 & 73.63 & 43.65 & $4.99 \times 10^{17}$ \\
MCTS                   & 60.12 & 10.80 & 22.77 & 56.80 & 40.71 & 74.74 & 44.32 & $9.33 \times 10^{17}$ \\
Guided Decoding        & 53.90 & 10.80 & 27.46 & 53.20 & 36.71 & 73.55 & 42.60 & $7.03 \times 10^{17}$ \\
Predictive Decoding    & 58.00 & 11.00 & 22.10 & 54.20 & 39.78 & 73.55 & 43.11 & $4.73 \times 10^{17}$ \\
\addlinespace[2pt]
$\phi$-Decoding        & 60.42 & 16.40 & 29.24 & 58.20 & 43.01 & 78.16 & 47.57 & $3.55 \times 10^{17}$ \\
Ours                   & \textbf{61.64} & \textbf{17.60} & \textbf{30.58} & \textbf{59.00} & \textbf{43.16} & \textbf{79.95} & \textbf{48.63} & $\mathbf{2.86 \times 10^{17}}$ \\   
\bottomrule
\end{tabularx}
\caption{Reasoning benchmark results of different decoding strategies. Best results for each model are highlighted in bold.}
\label{tab:main_results}
\end{table*}

\subsection{BenchMarks And Reported Metric}
To comprehensively evaluate the LLM performances on diverse tasks and fight head-to-head with some of the latest methods like Predictive Decoding \citep{manon} and $\phi$-decoding \citep{xu2025phi}. We evaluate on six representative reasoning benchmarks: GSM8K \citep{cobbe2021training}, MATH-500 \citep{hendrycks2021measuring}, GPQA \citep{rein2024gpqa}, ReClor \citep{yureclor}, LogiQA \citep{liu2021logiqa}, and ARC-Challenge \citep{clark2018think}. Together, these datasets span arithmetic word problems, competition-level mathematics, graduate-level factual reasoning, formal logic, and commonsense multiple-choice evaluation. We report the Pass@1 accuracy (Acc.) and FLOPS (metric for efficiency) for each benchmark as the evalutation metric. 
The computation formula for FLOPS is as follows \citep{kaplan2020scaling}: $\text{FLOPS} \approx 6nP$, where n represents the total number of output tokens and P is the number of LLM parameters. 

\subsection{Baselines and Backbone LLMs}
We benchmark $\phi$-Decoding against five representative reasoning paradigms: 1) \textbf{Auto-Regressive (CoT)}~\citep{wei2022chain} is standard chain-of-thought reasoning via auto-regressive decoding; 2) \textbf{Tree-of-Thought (ToT)} \citep{yao2023tree} constructs a reasoning tree where each node denotes a step; we adopt the breadth-first search (BFS) variant; 3) \textbf{Monte Carlo Tree Search (MCTS)} \citep{hao2023reasoning} builds a search tree with iterative expansion and backtracking, following the \emph{Reasoning as Planning (RaP)} framework; 4) \textbf{Guided Decoding} \citep{xie2023self} performs stochastic beam search with self-evaluation at each step; 5) \textbf{Predictive Decoding} \citep{manon} incorporates look-ahead strategies via model predictive control to reweight LLM distributions and mitigate myopic generation; 6) \textbf{$\phi$-Decoding} \citep{xu2025phi} estimates the step value based on the joint distribution derived from foresight paths. Both in-depth and in-width pruning strategies are introduced to alleviate the overthinking issue without requiring external auxiliary models.

For evaluation, we consider six widely used reasoning benchmarks. All baselines are implemented on two mid-scale instruction-tuned LLMs: \textbf{LLaMA3.1-8B-Instruct} \citep{grattafiori2024llama} and \textbf{Mistral-v0.3-7B-Instruct} \citep{jiang2023clip}. To validate our method in the latest model, we extend the experiments to \textbf{Qwen2.5-3B-Instruct} \citep{yang2025qwen3}.
\subsection{Experimental Environments And Basic Settings}
All experiments were performed on two NVIDIA RTX 3090 GPUs. For inference and sampling, we adopt vLLM (v0.9.1) in conjunction with PyTorch (v2.7.0) to ensure efficient large-scale decoding. Unless otherwise specified, the decoding temperature of the LLMs of the backbone is fixed at 0.7, a widely adopted setting that balances exploration and determinism among various reasoning tasks. For our martingale-based early stopping criterion, we set the pruning threshold at $10^{-6}$, so that once all candidate trajectories cease exhibiting an upward martingale trend, further sampling is terminated. This configuration provides a principled trade-off between computational efficiency and predictive stability. As $\lambda_1$ controls the aggressiveness of variance-based pruning, we additionally examine its effect on accuracy and FLOPs (Appendix~\ref{appendix:lambda_ablation}).

\subsection{Main Results}
Table~\ref{tab:main_results} presents the performance of our method against five representative decoding strategies on six reasoning benchmarks, using both LLaMA3.1-8B-Instruct and Mistral-v0.3-7B-Instruct as backbone models. The results clearly demonstrate that our principled approach consistently establishes a new state-of-the-art in both accuracy and computational efficiency. In LLaMA3.1-8B-Instruct, our method achieves a new state-of-the-art average accuracy of 60.21, representing a 34.1\% relative improvement over the standard Auto-Regressive (CoT) baseline. It also surpasses the strongest existing foresight sampling method, $\phi$-Decoding. This superior accuracy is achieved with remarkable efficiency; our method requires only $4.38\times10^{17}$ FLOPS, which makes it 31.9\% more efficient than $\phi$-decoding and over 35\% more efficient than MCTS. This dual advantage is consistent on Mistral-v0.3-7B-Instruct, where our method's average accuracy of 48.63 marks a 19.63\% relative gain over the CoT baseline. In terms of computational cost, our approach is the most efficient among all search-based strategies, requiring 19.4\% fewer FLOPS than $\phi$-Decoding and 42.7\% fewer than Tree-of-Thoughts. 

\subsection{Ablation Studies and Analysis}
\paragraph{Ablation Studies}

\begin{table}[t]
\centering
\begin{adjustbox}{width=\columnwidth}
\begin{tabular}{lccc}
\toprule
\textbf{Models} & \textbf{GSM8K} & \textbf{ReClor} & \textbf{FLOPs} \\
\midrule
\rowcolor[gray]{0.93}
\multicolumn{4}{l}{\textbf{LLaMA3.1-8B-Instruct}}\\
\addlinespace[2pt]
Ours & 87.64 & 65.20 & \num{3.70e17} \\
w/o martingale-based early-stop & 85.75 & 64.00 & \num{5.85e17} \\
w/o optional path selection & 86.36 & 61.40 & \num{4.12e18} \\
\bottomrule
\end{tabular}
\end{adjustbox}
\caption{Ablation study on \textbf{LLaMA3.1-8B-Instruct}. FLOPs are estimated using $6nP$~\citep{kaplan2020scaling}, with $P{=}8$B. Step beam size and number of rollouts per step are set to 8.}
\label{tab:llama-ablation}
\end{table}

To isolate the contribution of the core components of our framework, we conducted a critical ablation study focusing on our martingale-based early stopping mechanism, as it is central to achieving the balance between performance and computational efficiency. In Table~\ref{tab:llama-ablation}, we compare our full MFS framework against a variant where this principled stopping criterion is removed or optional path selection is removed, forcing the model to continue its foresight sampling process for a fixed, extended duration. The results clearly demonstrate the dual benefit of our approach. As expected, the early stop mechanism and optional path selection produce a substantial efficiency gain, reducing the required computational cost by \textbf{36.8\%} and more in terms of FLOPS. More importantly, the results show that removing this criterion also degrades accuracy on the GSM8K and ReClor benchmarks. This suggests that our principled stopping rule does more than just saving computation; it actively prevents the model from overthinking a problem, a phenomenon where continued, unguided exploration can introduce noise and lead the model away from an already-converged optimal solution. This study confirms that the adaptive stopping rule, guided by the Martingale Convergence Theorem, is essential for achieving both state-of-the-art accuracy and efficiency.

\paragraph{Probing Alternative Design Choices}
To better understand whether components of the foresight-sampling framework better than prior methods, we examine an alternative design choice at the core of candidate pruning: the step-valuation function.  
Prior work commonly relies on heuristic $\phi$-decoding scores to prune intermediate reasoning paths, but these heuristics lack a principled connection to the underlying stochastic decision process.  
We instead introduce a \textbf{step-value estimator}, a Doob-style advantage function $V(a_t)$ that predicts the utility of each partial reasoning trajectory.

To isolate the effect of valuation quality, we keep all other components, including beam size, rollout depth, stopping rule, and pruning thresholds fixed.  
As shown in Table~\ref{tab:stepvalue_vs_phi}, the step-value estimator consistently improves both accuracy and computational efficiency.  
On \textbf{ReClor}, accuracy increases from \textbf{64.00} to \textbf{65.20} with a \textbf{1.47$\times$} FLOPs reduction.  
On \textbf{MATH500}, accuracy improves from \textbf{37.20} to \textbf{38.20} alongside a \textbf{1.48$\times$} FLOPs reduction.  
These results demonstrate that a valuation signal grounded in advantage estimation provides a more reliable pruning criterion than heuristic scoring, enabling more effective allocation of computation toward promising reasoning paths.

\begin{table}[htbp]
\centering

\begin{adjustbox}{max width=\linewidth}
{
\setlength{\tabcolsep}{5pt}    % column padding
\renewcommand{\arraystretch}{1.15}  % row height

\begin{tabular}{lccc}
\toprule
\textbf{Scoring Method} &
\textbf{Acc} $\uparrow$ &
\textbf{FLOPs} ($\times 10^{17}$) $\downarrow$ &
\textbf{Speedup} \\
\midrule
\multicolumn{4}{c}{\textbf{ReClor}} \\
\cmidrule(lr){1-4}
Step-value (Ours) &
\textbf{65.20} &
\textbf{2.73} &
\textbf{1.47$\times$} \\
$\phi$-Decoding Score &
64.00 &
4.01 &
1$\times$ \\
\midrule
\multicolumn{4}{c}{\textbf{MATH500}} \\
\cmidrule(lr){1-4}
Step-value (Ours) &
\textbf{38.20} &
\textbf{10.22} &
\textbf{1.48$\times$} \\
$\phi$-Decoding Score &
37.20 &
15.15 &
1$\times$ \\
\bottomrule
\end{tabular}
}
\end{adjustbox}

\caption{Comparison of step-value vs.\ $\phi$-decoding scores for candidate pruning (Beam=8, $\lambda_1=0.8$).}
\label{tab:stepvalue_vs_phi}
\end{table}

\paragraph{Generalization experiments on Qwen2.5-3B-Instruct}
To further assess the generalization ability of our framework, we evaluate MFS on
the recent compact model \textbf{Qwen2.5-3B-Instruct}~\cite{qwen2.5}, which is known for its
strong reasoning capability. The results of two classic benchmarks, ReClor 
and ARC-c, are reported in Table~\ref{tab:generalization}. MFS yields substantial
gains over both the standard autoregressive baseline and the state-of-the-art
\textbf{$\phi$-Decoding}~\citep{xu2025phi}. Notably, MFS improves ReClor by
\textbf{+10.18} and ARC-c by \textbf{+7.58} over the baseline, highlighting its
robustness across distinct model architectures. The improvements become even more
pronounced as model quality increases, suggesting that MFS benefits more from
advanced reasoning capabilities. For completeness, Algorithms~\ref{alg:comparison1}
and~\ref{alg:comparison2} present side-by-side pseudocode for MFS and
$\phi$-Decoding.

\begin{table}[htbp]
\centering

\begin{adjustbox}{max width=\linewidth}
{
\small
\renewcommand{\arraystretch}{1.3} % Increased row height for stacked text

\begin{tabular}{lcc}
\toprule
\textbf{Method} &
% \textbf{GSM8K} &
\textbf{ReClor} &
\textbf{ARC-c} \\
\midrule
% Baseline Row
AR (CoT) &
% 78.62 &
53.60 &
77.47 \\
\midrule
% Comparison Rows
Predictive Decoding &
% \shortstack{84.08 \\ \scriptsize{(+5.46)}} &
\shortstack{60.00 \\ \scriptsize{(+6.40)}} &
\shortstack{78.69 \\ \scriptsize{(+1.22)}} \\
\addlinespace[4pt] % Extra space between rows for readability

Guided Decoding &
% \shortstack{87.41 \\ \scriptsize{(+8.79)}} &
\shortstack{60.40 \\ \scriptsize{(+6.80)}} &
\shortstack{78.34 \\ \scriptsize{(+0.87)}} \\
\addlinespace[4pt]

Tree-of-Thoughts &
% \shortstack{85.22 \\ \scriptsize{(+6.60)}} &
\shortstack{60.40 \\ \scriptsize{(+6.80)}} &
\shortstack{78.23 \\ \scriptsize{(+0.76)}} \\
\addlinespace[4pt]

$\phi$-Decoding &
% \shortstack{85.60 \\ \scriptsize{(+6.98)}} &
\shortstack{59.40 \\ \scriptsize{(+5.80)}} &
\shortstack{79.69 \\ \scriptsize{(+2.22)}} \\
\addlinespace[4pt]

\textbf{Ours} &
% \shortstack{\textbf{90.00} \\ \scriptsize{\textbf{(+11.38)}}} &
\shortstack{\textbf{63.78} \\ \scriptsize{\textbf{(+10.18)}}} &
\shortstack{\textbf{85.05} \\ \scriptsize{\textbf{(+7.58)}}} \\
\bottomrule
\end{tabular}
}
\end{adjustbox}

\caption{Generalization experiments on Qwen2.5-3B-Instruct. Values in parentheses denote improvement over the AR (CoT) baseline.}
\label{tab:generalization}
\end{table}

\section{Conclusion}
In summary, we introduced MFS, a principled framework that reframes LLM decoding from a heuristic search into the problem of identifying an optimal stochastic process. By moving beyond ad-hoc design choices, we demonstrated that the core components of an advanced decoder can be derived directly from foundational theorems in Martingale theory. Our approach models the quality of a reasoning path as a stochastic process with the objective of finding and extending paths that behave like submartingales, those whose value is expected to increase at every step. Experiments conducted on six reasoning benchmarks show that MFS surpasses state-of-the-art methods in accuracy while simultaneously improving computational efficiency.

\section{Limitations}
While MFS establishes a principled and theoretically grounded decoding framework, several limitations merit discussion.

First, our current evaluation focuses primarily on convergent reasoning tasks (e.g., mathematics, symbolic logic). This is inherent to the theoretical foundations of MFS: our formulation relies on the Martingale Convergence Theorem (Theorem 2), which guarantees stabilization toward a well-defined terminal value (the correct answer). Open-ended or creative generation does not provide such a limit, making martingale-style convergence inappropriate for optimizing diversity or stylistic quality.

Second, although MFS is far more computationally efficient than classical search-based methods, it still introduces additional overhead compared with standard autoregressive decoding due to the need to simulate future rollouts. Our adaptive stopping rule,derived from martingale convergence, substantially reduces redundant computation, but the inherent foresight cost remains a constraint.

Third, the effectiveness of MFS depends on the quality of the base model’s foresight probability $F_t$. The Doob decomposition used in MFS assumes a meaningful separation between predictable drift (reasoning signal) and the martingale residual (noise). If the underlying LLM is poorly calibrated or generates weak predictive signals, the advantage estimation and pruning decisions may degrade. Improving foresight calibration is therefore an important direction for future research.

Finally, our empirical results span six structured reasoning benchmarks. While these demonstrate strong gains, the behavior of MFS on long-form, multi-turn tasks remains unexplored. 

\bibliographystyle{acl_natbib}
\bibliography{custom}
\clearpage

\appendix
\label{sec:appendix}
\section{Detailed comparison versus $\phi$-decoding}
\algrenewcommand\algorithmicindent{1.0em}
\setlength{\emergencystretch}{1em} % 可选

% --- side-by-side algorithms spanning two columns ---
\begin{figure*}[t]
\centering
\begin{minipage}[t]{0.485\textwidth}
\raggedright\footnotesize
\begin{algorithmic}[1]\raggedright
\State \textbf{Input:} Query $x$, Initial beams $B_0$, Step beam size $M$, Num rollouts $N$
\State \textbf{Output:} Final reasoning path
\State \textbf{Initialize:} \texttt{beams} $\gets B_0$, $t \gets 0$
\While{\textbf{not} \texttt{convergence\_\allowbreak condition\_\allowbreak met}}
  \State \texttt{candidate\_paths} $\gets$ Generate $M \times N$ rollouts
  \State \Comment{---- Valuation Stage: The Core Difference ----}
  
  {\color{red}
  \Statex \textit{// For MFS (Principled Valuation)}
  \For{each path $p$ \textbf{in} \texttt{candidate\_paths}}
    \State \Comment{Estimate trend via Doob Decomposition}
    \State $F_t \gets$ Estimate future quality
    \State $V(p) \gets F_t - F_{t-1}$ \Comment{Predictable Advantage}
  \EndFor
  }
 
  {\color{blue}
 \Statex \textit{// For $\phi$-Decoding (Heuristic Valuation)}
  \For{each path $p$ \textbf{in} \texttt{candidate\_paths}}
    \State $F_t \gets$ Estimate future quality
    \State $\text{Advantage} \gets F_t - F_{t-1}$
    \State $\text{Alignment} \gets$ Score from clustering paths
    \State $R(p) \gets \text{Combine}(\text{Advantage}, \text{Alignment})$
  \EndFor
  }
\EndWhile
\end{algorithmic}
\captionof{algorithm}{Decoding Loop (Part 1)}
\label{alg:comparison1}
\end{minipage}\hfill
\begin{minipage}[t]{0.485\textwidth}
\raggedright\footnotesize
\begin{algorithmic}[1]\raggedright
\State \Comment{---- Path Selection Stage ----}

{\color{red}
\Statex \textit{// For MFS}
\State $\text{weights} \gets \text{softmax}([V(p) \text{ for } p])$
\State \texttt{beams} $\gets \text{sample}(M, \text{paths}, \text{weights})$
}

{\color{blue}
\Statex \textit{// For $\phi$-Decoding}
\State $\text{weights} \gets \text{softmax}([R(p) \text{ for } p])$
\State \texttt{beams} $\gets \text{sample}(M, \text{paths}, \text{weights})$
}
\State \Comment{---- Convergence Check ----}
{\color{red}
\Statex \textit{// For MFS (Martingale Convergence)}
\If{$\max(V(p)) \le \epsilon_{\text{stop}}$}
  \State \texttt{convergence\_\allowbreak condition\_\allowbreak met} $\gets$ \textbf{True}
\EndIf
}
{\color{blue}
\Statex \textit{// For $\phi$-Decoding (Consensus)}
\If{$\text{max\_cluster\_size} / \text{num\_paths} \ge \delta$}
  \State \texttt{convergence\_\allowbreak condition\_\allowbreak met} $\gets$ \textbf{True}
\EndIf
}
\State $t \gets t + 1$
\State \textbf{Return} best path from \texttt{beams}
\end{algorithmic}
\captionof{algorithm}{Decoding Loop (Part 2)}
\label{alg:comparison2}
\end{minipage}
\end{figure*}

\hfill

This section compares the proposed MFS and $\phi$-decoding in detail as shown in Algorithms \ref{alg:comparison1} and \ref{alg:comparison2}. This comparison demonstrates the fundamental architectural difference: MFS replaces the \textbf{heuristic-based components} of prior work with \textbf{principled mechanisms} derived directly from Martingale theory. Specifically, the algorithms highlight how our method substitutes the ad-hoc \textit{Advantage + Alignment} scoring with a theoretically grounded valuation based on the \textit{Doob Decomposition} and replaces the consensus-based stopping rule with an adaptive criterion guided by the \textit{Martingale Convergence Theorem}. This illustrates how our principled design directly translates to superior empirical performance and efficiency.

\section{Parameters}
\newcolumntype{Y}{>{\raggedright\arraybackslash}X}

% 定义超参数展示命令
\newcommand{\hpsimple}[3]{%
  \hfill $M=#1$, $N=#2$, $\lambda_1=#3$ \hfill
}

\begin{table}[!t]
  \centering
  \small
  \renewcommand{\arraystretch}{1.15}
  \setlength{\tabcolsep}{6pt}
    \begin{tabularx}{\linewidth}{lX}

    \toprule
    \textbf{Task} & \textbf{Hyper-Parameter Setup} \\
    \midrule

    % ===== LLaMA3.1-8B =====
    \rowcolor{gray!12}
    \multicolumn{2}{c}{\textbf{LLaMA3.1-8B-Instruct}} \\
    GSM8K    & \hpsimple{8}{8}{0.8} \\
    MATH-500 & \hpsimple{8}{8}{0.6} \\
    GPQA     & \hpsimple{8}{8}{0.6} \\
    ReClor   & \hpsimple{8}{8}{0.8} \\
    LogiQA   & \hpsimple{8}{8}{0.6} \\
    ARC-C    & \hpsimple{8}{8}{0.8} \\

    % ===== Mistral =====
    \rowcolor{gray!12}
    \multicolumn{2}{c}{\textbf{Mistralv0.3-7B-Instruct}} \\
    GSM8K    & \hpsimple{8}{8}{0.6} \\
    MATH-500 & \hpsimple{8}{8}{0.8} \\
    GPQA     & \hpsimple{8}{8}{0.8} \\
    ReClor   & \hpsimple{8}{8}{0.6} \\
    LogiQA   & \hpsimple{8}{8}{1.0} \\
    ARC-C    & \hpsimple{8}{8}{1.0} \\

    % ===== Qwen =====
    \rowcolor{gray!12}
    \multicolumn{2}{c}{\textbf{Qwen2.5-3B-Instruct}} \\
    ReClor   & \hpsimple{8}{8}{0.8} \\
    ARC-C    & \hpsimple{8}{8}{0.8} \\

    \bottomrule
  \end{tabularx}
  \caption{Experimental setup of our martingale-principled decoding. $M$ denotes the step beam size, $N$ is the number of rollouts per step beam, and $\lambda_1$ is the parameter for principled pruning.}
  \label{tab:setup}
\end{table}

This section details the hyperparameter configurations used for our MFS experiments. The complete settings for each task and the backbone model are presented in Table \ref{tab:setup}. Our framework is notably robust, requiring minimal task-specific tuning. For all experiments across all models, we kept the core search parameters constant: The size of the step beam ($M$), was set to $8$, and the number of rolls per step beam ($N$) was also set to $8$. The primary hyperparameter we tuned was $\lambda_1$,  which controls the sensitivity of our principal pruning mechanism. As shown in the table, this parameter was adjusted within a narrow range (0.6 to 1.0) to optimize performance for specific model-task pairs.

\section{Analysis of the Pruning Coefficient $\lambda_1$}
\label{appendix:lambda_ablation}

The coefficient $\lambda_1$ controls the aggressiveness of our variance-based
pruning rule (Eq.~11). Lower $\lambda_1$ retains more candidate trajectories,
leading to higher computational cost but stable accuracy; higher $\lambda_1$
tightens the pruning threshold, removing noisy trajectories earlier.

Table~\ref{tab:lambda_ablation_appendix} reports the accuracy--efficiency trade-off
across ReClor and MATH500 under different $\lambda_1$ and beam sizes. We observe
that $\lambda_1 = 0.8$ achieves the best balance, reducing FLOPs by approximately
20\% while slightly improving accuracy. This supports using a moderately strict
variance threshold to prevent wasted compute on unstable reasoning paths.

\begin{table}[t]
\centering
\small
\resizebox{1\columnwidth}{!}{
\begin{threeparttable}
\setlength{\tabcolsep}{6pt}
\begin{tabular}{
c c 
S[table-format=2.1] 
S[table-format=1.2] 
S[table-format=2.1] 
S[table-format=2.2]
}
\toprule
{$\lambda_1$} & {$b$} &
{\textbf{ReClor Acc}~($\uparrow$)} &
{\textbf{ReClor FLOPs}~($\downarrow$)} &
{\textbf{MATH500 Acc}~($\uparrow$)} &
{\textbf{MATH500 FLOPs}~($\downarrow$)} \\
\midrule
0.6 & 4 & 61.0 & 0.70 & 35.0 & 2.72 \\
0.6 & 6 & 62.4 & 1.81 & 37.0 & 5.65 \\
0.6 & 8 & 64.0 & 3.42 & \bfseries 38.2 & 10.22 \\
\addlinespace
0.8 & 4 & 61.8 & 0.62 & 33.8 & 1.80 \\
0.8 & 6 & 64.3 & 1.52 & 34.4 & 4.19 \\
0.8 & 8 & \bfseries 65.2 & \bfseries 2.73 & 37.2 & 6.92 \\
\bottomrule
\end{tabular}

\caption{Ablation on pruning coefficient $\lambda_1$ and beam size $b$ 
for LLaMA3.1-8B-Instruct on ReClor and MATH500. 
$\uparrow$ higher is better, $\downarrow$ lower is better.}
\label{tab:lambda_ablation_appendix}
\end{threeparttable}
}
\end{table}
\end{document}